\documentclass[runningheads]{llncs}
\usepackage{cite}
\usepackage{amsmath,amssymb,amsfonts}
\usepackage{algorithmic}
\usepackage{graphicx}
\usepackage{textcomp}
\usepackage{xcolor}
\usepackage{float}
\usepackage{hyperref}
\usepackage{multirow}
\usepackage{svg}
\usepackage{adjustbox}
\usepackage{siunitx}
\def\BibTeX{{\rm B\kern-.05em{\sc i\kern-.025em b}\kern-.08em
    T\kern-.1667em\lower.7ex\hbox{E}\kern-.125emX}}
\begin{document}
\title{YOLOPoint: Joint Keypoint and Object Detection\thanks{This research is funded by dtec.bw -- Digitalization and Technology Research Center of the Bundeswehr. dtec.bw is funded by the European Union - NextGenerationEU.}}
\author{Anton Backhaus \and Thorsten Luettel \and Hans-Joachim Wuensche}%
\authorrunning{A. Backhaus et al.}
\institute{Institute of Autonomous Systems Technology, University of the Bundeswehr Munich
\email{anton.backhaus@unibw.de}\\}
\maketitle              
\begin{abstract}
Intelligent vehicles of the future must be capable of understanding and navigating safely through their surroundings.
Camera-based vehicle systems can use keypoints as well as objects as low- and high-level landmarks for GNSS-independent SLAM and visual odometry.
To this end we propose YOLOPoint, a convolutional neural network model that simultaneously detects keypoints and objects in an image by combining YOLOv5 and SuperPoint to create a single forward-pass network that is both real-time capable and accurate.
By using a shared backbone and a light-weight network structure, YOLOPoint is able to perform competitively on both the HPatches and KITTI benchmarks.

\keywords{Deep learning  \and Keypoint detection \and Autonomous driving}
\end{abstract}
\section{Introduction}\label{section:Introduction}
Keypoints are low-level landmarks, typically points, corners, or edges that can easily be retrieved from different viewpoints.
This makes it possible for moving vehicles to estimate their position and orientation relative to their surroundings and even perform loop closure (i.e., SLAM) with one or more cameras.
Historically, this task was performed with hand-crafted keypoint feature descriptors such as ORB~\cite{bib:orb}, SURF~\cite{bib:surf}, HOG~\cite{bib:hog}, SIFT~\cite{bib:sift}.
However, these are either not real-time capable or perform poorly under disturbances such as illumination changes, motion blur, or they detect keypoints in clusters rather than spread out over the image, making pose estimation less accurate.
Learned feature descriptors aim to tackle these problems, often by applying data augmentation in the form of random brightness, blurring and contrast.
Furthermore, learned keypoint descriptors have shown to outperform classical descriptors.
One such keypoint descriptor is SuperPoint~\cite{bib:superpoint}, a convolutional neural network (CNN) which we use as a base network to improve upon.

SuperPoint is a multi-task network that jointly predicts keypoints and their respective descriptors in a single forward pass.
It does this by sharing the feature outputs of one backbone between a keypoint detector and descriptor head.
This makes it computationally efficient and hence ideal for real-time applications.

Furthermore, after making various adjustments to the SuperPoint architecture, we fuse it with YOLOv5~\cite{bib:yolov5}, a real-time object detection network.
The full network thus uses one shared backbone for all three tasks.
Moreover, using the YOLOv5 framework we train four models of various sizes: \textit{nano}, \textit{small}, \textit{middle} and \textit{large}.
We call the combined network \textit{YOLOPoint}.

Our motivation for fusing the keypoint detector with the object detector is as follows: 
Firstly, for a more accurate SLAM and a better overview of the surrounding scene, using a suite of multi-directional cameras is beneficial. 
But processing multiple video streams in parallel means relying on efficient CNNs.
In terms of computational efficiency, the feature extraction part does most of the heavy lifting.
This is why it has become common practice for multiple different tasks to share a backbone.
SuperPoint’s architecture is already such that is uses a shared backbone for the keypoint detection and description, hence it was an adequate choice for fusion with another model.
Secondly, visual SLAM works on the assumption of a static environment. 
If keypoints are detected on moving objects, it would lead to localization errors and mapping of unwanted landmarks.
However, with classified object bounding boxes we simply filter out all keypoints within a box of a dynamic class (e.g., pedestrian, car, cyclist).
Finally, although it seems that keypoint and object detection are too different to jointly learn, they have been used in conjunction in classical methods (e.g., object detection based on keypoint descriptors with a support vector machine classifier \cite{bib:objectclassical}).

\begin{figure}[t]   
	\begin{center}
		\includegraphics[trim={18mm 0mm 42mm 0mm},clip,width=1.\linewidth]{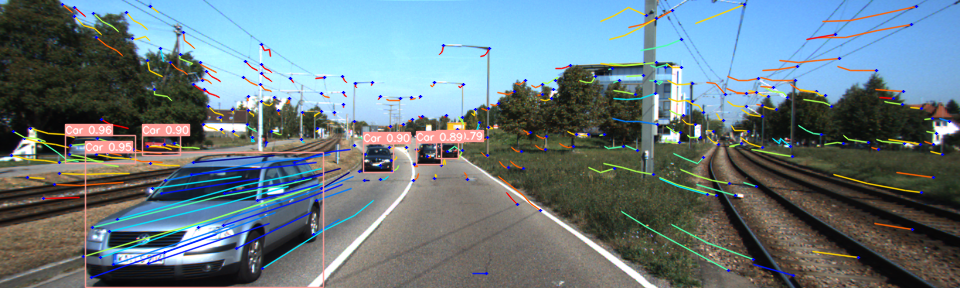}
		\caption{Example output of YOLOPointM on a KITTI scene with keypoint tracks from 3 frames and object bounding boxes.}
		\label{fig:map}
	\end{center}
	\vspace*{-7pt}
\end{figure}

Our contributions can be summarized as follows:
\begin{itemize}
\item We propose YOLOPoint, a fast and efficient keypoint detector and descriptor network which is particularly well suited for performing visual odometry.
\item We show that object and keypoint detection are not mutually exclusive and propose a network that can do both in a single forward pass.
\item We demonstrate the efficacy of using efficient cross stage partial network (CSP) blocks \cite{bib:cspdarknet} for point description and detection tasks.
\end{itemize}
Code will be made available at \url{https://github.com/unibwtas/yolopoint}.

\section{Related Work}\label{Related Work}

Classical keypoint descriptors use hand-crafted algorithms designed to overcome challenges such as scale and illumination variation~\cite{bib:fast, bib:orb, bib:surf, bib:sift, bib:hog, bib:schweitzer2009ecvw} and have been thoroughly evaluated~\cite{bib:performanceclassical, bib:performanceclassical2}.
Although their main utility was keypoint description, they have also been used in combination with support vector machines to detect objects~\cite{bib:objectclassical}.

Since then, deep learning-based methods have dominated benchmarks for object detection (i.e., object localization and classification) and other computer vision tasks \cite{bib:objectclassical, bib:object_detection_review_2, bib:object_detection_review_3}.
Therefore, research has increasingly been dedicated to finding ways in which they can also be used for point description.
Both using CNN-based transformer models COTR~\cite{bib:cotr} and LoFTR~\cite{bib:loftr} achieve state-of-the-art results on multiple image matching benchmarks.
COTR finds dense correspondences between two images by concatenating their respective feature maps and processing them together with positional encodings in a transformer module.
Their method, however, focuses on matching accuracy rather than speed and performs inference on multiple zooms of the same image.
LoFTR has a similar approach, the main difference being their "coarse-to-fine" module that first predicts coarse correspondences, then refines them using a crop from a higher level feature map.
Both methods are detector-free, and while achieving excellent results in terms of matching accuracy, neither method is suitable for real-time applications.
Methods that yield both keypoint detections and descriptors are generally faster due to matching only a sparse set of points and include R2D2~\cite{bib:r2d2}, D2-Net~\cite{bib:d2net} and SuperPoint~\cite{bib:superpoint}.
R2D2 tackles the matching problems of repetitive textures by introducing a reliability prediction head that indicates the discriminativeness of a descriptor at a given pixel.
D2-Net has the unique approach of using the output feature map both for point detection and description, hence sharing all the weights of the network between both tasks.
In contrast, SuperPoint has a shared backbone but seperate heads for the detection and description task.
What sets it apart from all other projects is its self-supervised training framework.
While other authors obtain ground truth point correspondences from depth images gained from structure from motion, i.e., using video, the SuperPoint framework can create labels for single images.
It does this by first generating a series of labeled grayscale images depicting random shapes, then training an intermediate model on this synthetic data.
The intermediate model subsequently makes point-predictions on a large data set (here: MS COCO~\cite{bib:coco}) that are refined using their "homographic adaptation" method.
The final model is trained on the refined point labels.

While there exist several models that jointly predict keypoints and descriptors, there are to our knowledge none that also detect objects in the same network.
Maji et al's work~\cite{bib:yolopose} comes closest to ours.
They use YOLOv5 to jointly predict keypoints for human pose estimation as well as bounding boxes in a single forward pass.
The main differences are that the keypoint detection training uses hand-labelled ground truth points, the object detector is trained on a single class (human), and both tasks rely on similar features.

\section{Model Architecture}\label{section:Model Architecture}
\begin{figure*}[h!]
	\makebox[\textwidth][c]{\includegraphics[trim={0mm 58mm 30mm 1mm},clip ,width=01\textwidth]{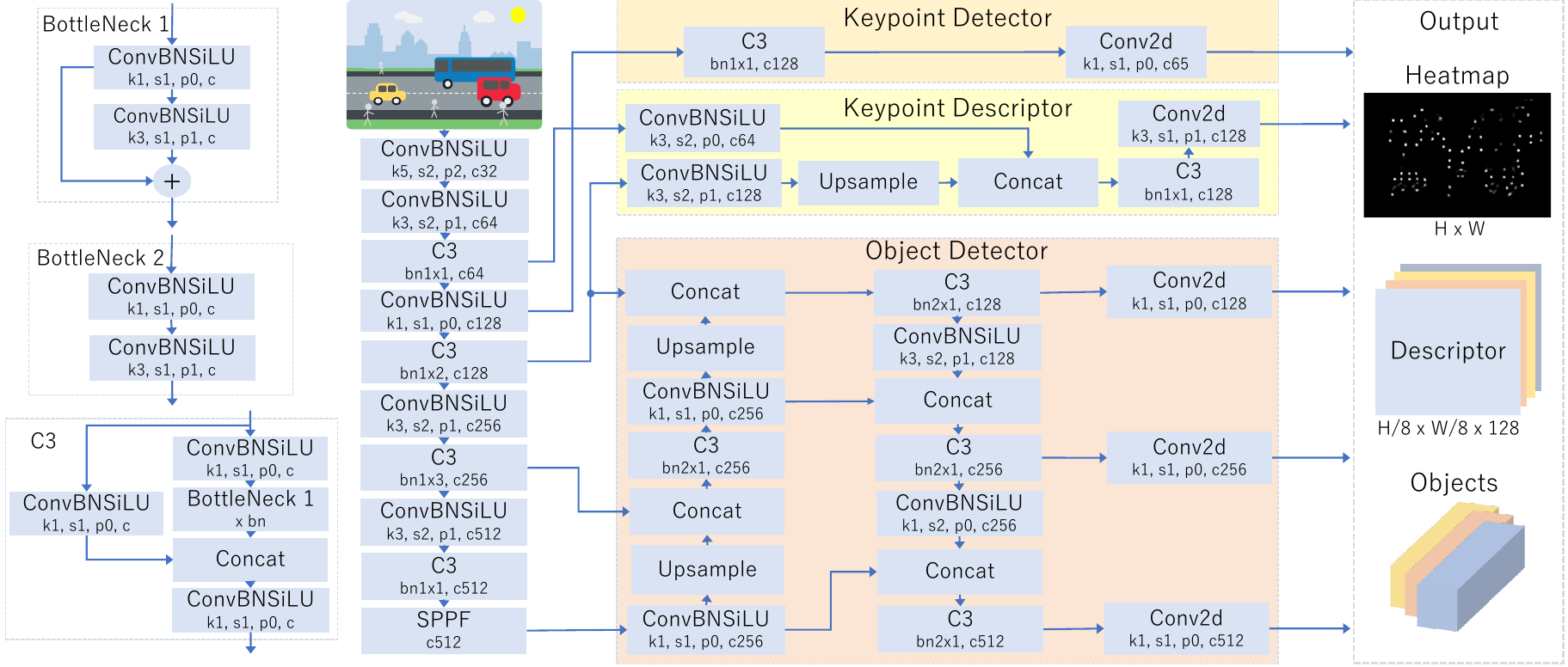}}
	\caption{
            Full model architecture exemplary for YOLOPointS.
            The two types of bottlenecks, C3 block (left) and a sequence of convolution, batch normalization and SiLU activation form the main parts of YOLO and by extension YOLOPoint.
            $k$: kernel size, $s$: stride, $p$: pad, $c$: output channels, $bn$: bottleneck, \emph{SPPF}: fast spacial pyramid pooling \cite{bib:yolov5}.}
	\label{fig:architecture}
\end{figure*}
Our keypoint detection model is an adaptation of SuperPoint with CSPDarknet~\cite{bib:cspdarknet} elements.
SuperPoint uses a VGG-like encoder~\cite{bib:vgg} that breaks down a grayscale image of size $H \times W \times 1$ into a feature vector of size $H/8 \times W/8 \times 128$.
In each decoder head the feature vector is further processed and reshaped to the original image size.
The keypoint decoder branch thus produces a heat map representing the pixel-wise probability of "point-ness", which is passed through a non-maximum suppression method that produces the final keypoints.
The output of the descriptor branch is a normed descriptor vector linearly up-scaled to $H \times W \times 256$.
Both outputs are combined into a list of keypoints and corresponding descriptor vectors, with which keypoints can be matched from frame to frame.

Our version of SuperPoint substitutes the VGG-like backbone with part of CSPDarknet and has a CSP bottleneck with additional convolutions in each head. 
CSP bottlenecks are the cornerstone of YOLO and provide a good speed to accuracy trade-off for various different computer vision tasks but have so far not been used in keypoint detection for visual landmarks.
We keep the same final layers as SuperPoint, i.e., softmax and reshape in the detector head and 2D convolution in the descriptor head.
Using YOLOv5's scaling method we create four networks, each varying in width, depth and descriptor length.
The descriptors $\mathcal{D}$ are sized at 256, 196, 128 and 64 for models YOLOPointL, -M, -S and -N respectively.
A shorter descriptor reduces the computational cost of matching at the cost of accuracy.
For comparison: SIFT’s descriptor vector has 128 elements and ORB’s has only 32.
Since reducing the vector size comes at an accuracy trade-off, the smallest vector is left at 64.
Furthermore, in order for the descriptor to be able to match and distinguish between other keypoints, it requires a large receptive field \cite{bib:loftr}.
This, however, comes with down-sampling the input image and loosing detail in the process.
In order to preserve detail, we enlarge the low-resolution feature map with nearest-neighbor interpolation and concatenate it with a feature map higher up in the backbone before performing further convolutions.
The full model fused with YOLOv5 is depicted in fig. \ref{fig:architecture}.

\section{Training}\label{section:Training}

To generate pseudo ground truth point labels, we follow the protocol of SuperPoint by first training the point detector of YOLOPoint on the synthetic shapes dataset, then using it to generate refined outputs on COCO using homographic adaptation for pre-training.
Pre-training on COCO is not strictly necessary, however it can improve results when fine-tuning on smaller data sets, as well as reduce training time.
Thus, the pre-trained weights are later fine-tuned on the KITTI dataset \cite{bib:kitti}

For training the full model, pairs of RGB images that are warped by a known homography are each run through a separate forward pass. 
The model subsequently predicts “point-ness” heat maps, descriptor vectors and object bounding boxes.
When training on data sets of variable image size (e.g., MS COCO) the images must be fit to a fixed size in order to be processed as a batch.
A common solution is to pad the sides of the image such that $W = H$, also known as letterboxing.
However, we found that this causes false positive keypoints to be predicted close to the padding due to the strong contrast between the black padding and the image, negatively impacting training.
Therefore, when pre-training on COCO, we use mosaic augmentation, that concatenates four images side-by-side to fill out the entire image canvas, eliminating the need for image padding \cite{bib:yolov4}.
All training is done using a batch size of 64 and the Adam optimizer \cite{bib:adam} with a learning rate of $10^{-3}$ for pre-training and $10^{-4}$ for fine-tuning.

For fine-tuning on KITTI we split the data into 6481 training and 1000 validation images resized to $288 \times 960$.
To accommodate the new object classes we replace the final object detection layer and train for 20 epochs with all weights frozen except those of the detection layer.
Finally, we unfreeze all layers and train the entire network for another 50 epochs.

\subsection{Loss Functions}\label{section:loss}
The model outputs for the warped and unwarped image are used to calculate the keypoint detector and descriptor losses. 
However, only the output of the unwarped images is used for the object detector loss, as to not teach strongly distorted object representations.

The keypoint detector loss $\mathcal{L}_{\text{det}}$ is the mean of the binary cross-entropy losses over all pixels of the heatmaps of the warped and unwarped image of size $H \times W$ and corresponding ground truth labels and can be expressed as follows:

\begin{equation}
\mathcal{L}_{\text{det}}=-\frac{1}{HW}\sum_{}^{H,W}(y_{ij} \cdot \log x_{ij}+(1-y_{ij}) \cdot \log (1-x_{ij}))
\end{equation}

where $y_{ij}\in\{0,1\}$ and $x_{ij}\in[0,1]$ respectively denote the target and prediction at pixel $ij$.

The original descriptor loss is a contrastive hinge loss applied to all correspondences and non-correspondences of a low-resolution descriptor $\mathcal{D}$ of size $H_c \times W_c$ \cite{bib:superpoint} creating a total of $(H_c \times W_c)^{2}$ matches.
This, however, becomes computationally unfeasible for higher resolution images.
Instead, we opt for a sparse version adapted from DeepFEPE \cite{bib:deepfepe} that samples $N$ matching pairs of feature vectors $d_{ijk}$ and $d_{i^{\prime}j^{\prime}k}$ and $M$ non-matching pairs of a batch of sampled descriptors $\Tilde{\mathcal{D}}\subset\mathcal{D}$ and their warped counterpart $\Tilde{\mathcal{D}^{\prime}}\subset\mathcal{D}^{\prime}$.
Using the known homography, each descriptor $\mathbf{d}$ of pixel $ij$ of the $k${th} image of a mini-batch can be mapped to its corresponding warped pair at $i^{\prime}j^{\prime}$.
$m_p$ furthermore denotes the positive margin of the sampled hinge loss.

\begin{equation}
\mathcal{L}_{\text{desc}}=\mathcal{L}_{\text{corr}}+\mathcal{L}_{\text{n.corr}}
\end{equation}
where
\begin{equation}
\mathcal{L}_{corr}=\frac{1}{N} \sum_{i,j,k,i^{\prime},j^{\prime}}^{\Tilde{\mathcal{D}},\Tilde{\mathcal{D^{\prime}}}} \max(0, m_p-\mathbf{d}_{ijk}^T \mathbf{d}_{i^{\prime}j^{\prime} k}^{\prime})
\end{equation}
\begin{equation}
\mathcal{L}_{n.corr}=\frac{1}{M} \sum_{i,j,k}^{\Tilde{\mathcal{D}}}\sum_{o,p,q}^{\Tilde{\mathcal{D^{\prime}}}}(\mathbf{d}_{ijk}^T \mathbf{d^{\prime}}_{opq}) ,\quad (o,p,q) \neq (i^{\prime},j^{\prime},k)
\end{equation}

The object detector loss $\mathcal{L}_{\text{obj}}$ is a linear combination of intermittent losses based on the objectness, class probability and bounding box regression score and is identical to the loss function used in YOLOv5.

The final loss is calculated as the weighted sum of the keypoint detector, descriptor and object loss.

\begin{equation}
	\mathcal{L} = \mathcal{L}_{\text{det}} + \mathcal{L}_{\text{det,warp}} + w_{\text{desc}} \cdot \mathcal{L}_{\text{desc}} + w_{\text{obj}} \cdot \mathcal{L}_{\text{obj}}
\end{equation}

\section{Evaluation}\label{section:Evaluation}

\begin{figure}[t]
	\begin{center}
		\includegraphics[width=1.\linewidth]{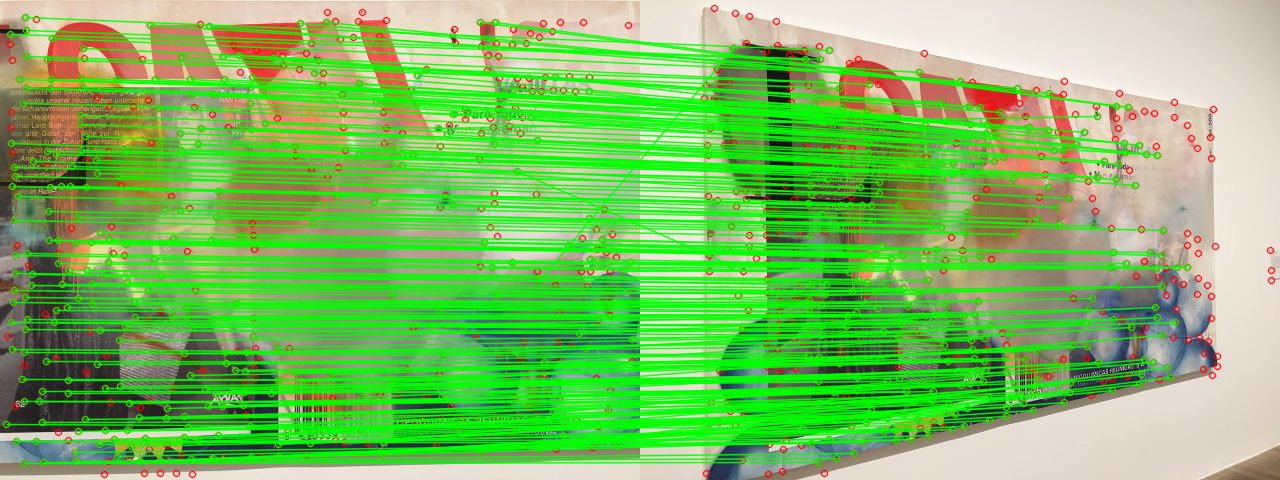}
		\caption{HPatches matches between two images with viewpoint change estimated with YOLOPointS. Matched keypoints are used to estimate the homography matrix describing the viewpoint change.}
		\label{fig:HPatches matches}
	\end{center}
\end{figure}

In the following sections we present our evaluation results for keypoint detection and description on HPatches \cite{bib:hpatches} and using all three task heads for visual odometry (VO) estimation on the KITTI benchmark.
For evaluation on HPatches the models trained for 100 epochs on MS COCO are used, for VO the models are fine-tuned on KITTI data.

\subsection{Repeatability and Matching on HPatches}\label{HPatches}
The HPatches dataset comprises a total of 116 scenes, each containing 6 images.
57 scenes exhibit large illumination changes and 59 scenes large viewpoint changes.
The two main metrics used for evaluating keypoint tasks are repeatability which quantifies the keypoint detector's ability to consistently locate keypoints at the same locations despite illumination and/or viewpoint changes and homography estimation which tests both repeatability and discrimination ability of the detector \emph{and} descriptor.
Our evaluation protocols are kept consistent with SuperPoint's where possible.

\subsubsection{Repeatability}
Repeatability is computed at $256 \times 320$ resolution with up to 300 points detected per image using a non-maximum suppression of 8 pixels.
A keypoint counts as repeated if it is detected within $\epsilon = 3$ pixels in both frames.
The repeatability score determines the ratio of repeated keypoints to overall detected keypoints \cite{bib:superpoint}.
Table \ref{table:repeatability} summarizes the repeatability scores under viewpoint and illumination changes.
SuperPoint outperforms YOLOPoint in both illumination and viewpoint changes by a small margin.
Surprisingly, all four versions of YOLOPoint perform virtually the same, despite varying strongly in number of parameters.
A likely explanation for this is that the keypoint detector branch has more parameters than it actually needs with regard to the simplicity of the task.

\begin{table}[ht!]
\caption{
    Repeatability score on HPatches in scenes of strong illumination and viewpoint changes.
    SuperPoint outperforms YOLOPoint by a small margin in both scenarios. Best results are boldfaced.
    }
\begin{center}
\begin{tabular}{ccc}
\hline
\multirow{2}{*}{} & \multicolumn{2}{c}{Repeatability}                 \\ \cline{2-3} 
                  & \multicolumn{1}{c}{Illumination}  & Viewpoint     \\ \hline
YOLOPointL        & \multicolumn{1}{c}{.590}          & .540          \\ \hline
YOLOPointM        & \multicolumn{1}{c}{.590}          & .540          \\ \hline
YOLOPointS        & \multicolumn{1}{c}{.590}          & .540          \\ \hline
YOLOPointN        & \multicolumn{1}{c}{.589}          & .529         \\ \hline
SuperPoint        & \multicolumn{1}{c}{\textbf{.611}} & \textbf{.555} \\ \hline
\end{tabular}
\end{center}
\vspace{-3mm}
\label{table:repeatability}
\end{table}

\subsubsection{Homography Estimation}
Homography estimation is computed at $320 \times 480$ resolution with up to 1000 points detected per image using a non-maximum suppression of 8 pixels.
By using matched points  between two frames (cf. fig. \ref{fig:HPatches matches}) a homography matrix that describes the transformation of points between both frames is estimated.
The estimated homography is then used to transform the corners of one image to another.
A homography counts as correct if the l2 norm of the distance of the corners lies within a margin $\epsilon$ \cite{bib:superpoint}.

Table \ref{table:homography} shows the results of the homography estimation on the viewpoint variation scenes.
Overall, both models are roughly on par, although SuperPoint yields slightly better results within larger error margins $\epsilon$.
SuperPoint's descriptors have a significantly higher nearest neighbor mean average precision (NN mAP), indicating good discriminatory ability, however a worse matching score than YOLOPoint.
The matching score is a measure for both detector and descriptor, as it measures the ratio of ground truth correspondences over the number of proposed features within a shared viewpoint region \cite{bib:superpoint}.

\begingroup
\setlength{\tabcolsep}{6pt} 
\begin{table}[h]
\caption{Homography estimation on HPatches.
YOLOPoint and SuperPoint perform comparably to each other.
Additional descriptor metrics are included.
Best results are boldfaced.}
\begin{center}
\begin{tabular}{ccccll}
\hline
\multirow{2}{*}{} & \multicolumn{3}{c}{\begin{tabular}[c]{@{}c@{}}Homography\\ estimation\end{tabular}}    & \multicolumn{2}{c}{\begin{tabular}[c]{@{}c@{}}Descriptor\\ metrics\end{tabular}}              \\ \cline{2-6} 
                  & \multicolumn{1}{c}{$\epsilon=1$}         & \multicolumn{1}{c}{$\epsilon=3$}         & $\epsilon=5$         & \multicolumn{1}{c}{NN mAP}        & \begin{tabular}[c]{@{}c@{}}Matching\\ Score\end{tabular} \\ \hline
YOLOPointL        & \multicolumn{1}{c}{.390}          & \multicolumn{1}{c}{.739}          & .841          & \multicolumn{1}{c}{.650}          & \hspace{3mm}\textbf{.459}                                            \\ \hline
YOLOPointM        & \multicolumn{1}{c}{.410}          & \multicolumn{1}{c}{.712}          & .817          & \multicolumn{1}{c}{.625}          & \hspace{3mm}.449                                                     \\ \hline
YOLOPointS        & \multicolumn{1}{c}{\textbf{.420}} & \multicolumn{1}{c}{.729}          & .817          & \multicolumn{1}{c}{.592}          & \hspace{3mm}.436                                                     \\ \hline
YOLOPointN        & \multicolumn{1}{c}{.292}          & \multicolumn{1}{c}{.668}          & .803          & \multicolumn{1}{c}{.547}          & \hspace{3mm}.389                                                     \\ \hline
SuperPoint        & \multicolumn{1}{c}{.312}          & \multicolumn{1}{c}{\textbf{.742}} & \textbf{.851} & \multicolumn{1}{c}{\textbf{.756}} & \hspace{3mm}.409                                                     \\ \hline

\end{tabular}
\end{center}
\vspace{-3mm}
\label{table:homography}
\end{table}
\endgroup

\subsection{Visual Odometry Estimation on KITTI}

\begin{figure}[t]
	\begin{center}
		{\includegraphics[trim={30mm 2mm 30mm 10mm},clip,width=\columnwidth]{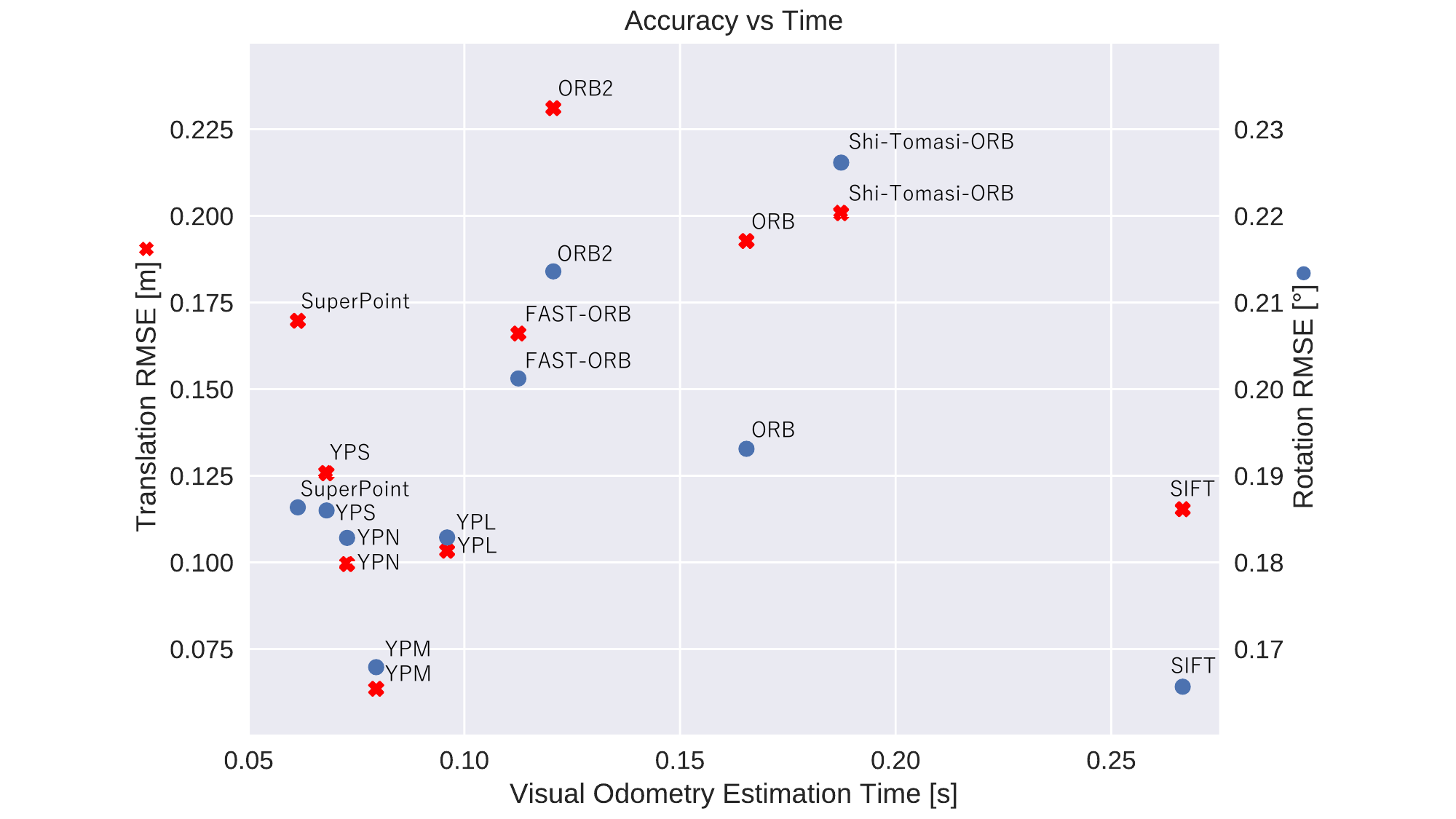}}
		\caption{Translation and rotation RMSE over all KITTI sequences plotted against mean VO estimation time for YOLOPointL (YPL), M, S and N with filtered points as well as SuperPoint and classical methods for comparison (lower left is better). VO estimation was done with $376 \times 1241$ images, NVIDIA RTX A4000 and Intel Core i7-11700K.}
		\label{fig:accuracy vs time}
	\end{center}
    \vspace{-2.7mm}
\end{figure}

The KITTI odometry benchmark contains 11 image sequences of traffic scenes with publicly available ground truth camera trajectories captured from a moving vehicle from the ego perspective.

Odometry is estimated using only frame-to-frame keypoint tracking (i.e., no loop closure, bundle adjustment, etc.), in order to get an undistorted evaluation of the keypoint detectors/descriptors.
In our tests we evaluate different versions of YOLOPoint with filtering out keypoints on dynamic objects using object bounding boxes and compare them to SuperPoint and other real-time classical methods.
Fig. \ref{fig:accuracy vs time} plots the RMSE over all sequences against the mean VO estimation time (i.e., keypoint detection, matching and pose estimation) for several models.
Despite the overhead of detecting objects, YOLOPoint provides a good speed-to-accuracy trade-off with a full iteration taking less than \SI{100}{\milli\second}.
The pure inference times of YOLOPoint (i.e. without pose estimation) are 49, 36, 27 and \SI{25}{\milli\second} for L, M, S and N, respectively.
Note that the VO time using YOLOPointN is longer despite having a faster inference since its keypoints have more outliers and noisier correspondences. Consequently, the pose estimation algorithm needs more iteration steps to converge, resulting in a slower VO.

\begin{figure}[t]
	\begin{center}
		\includegraphics[width=1.\linewidth]{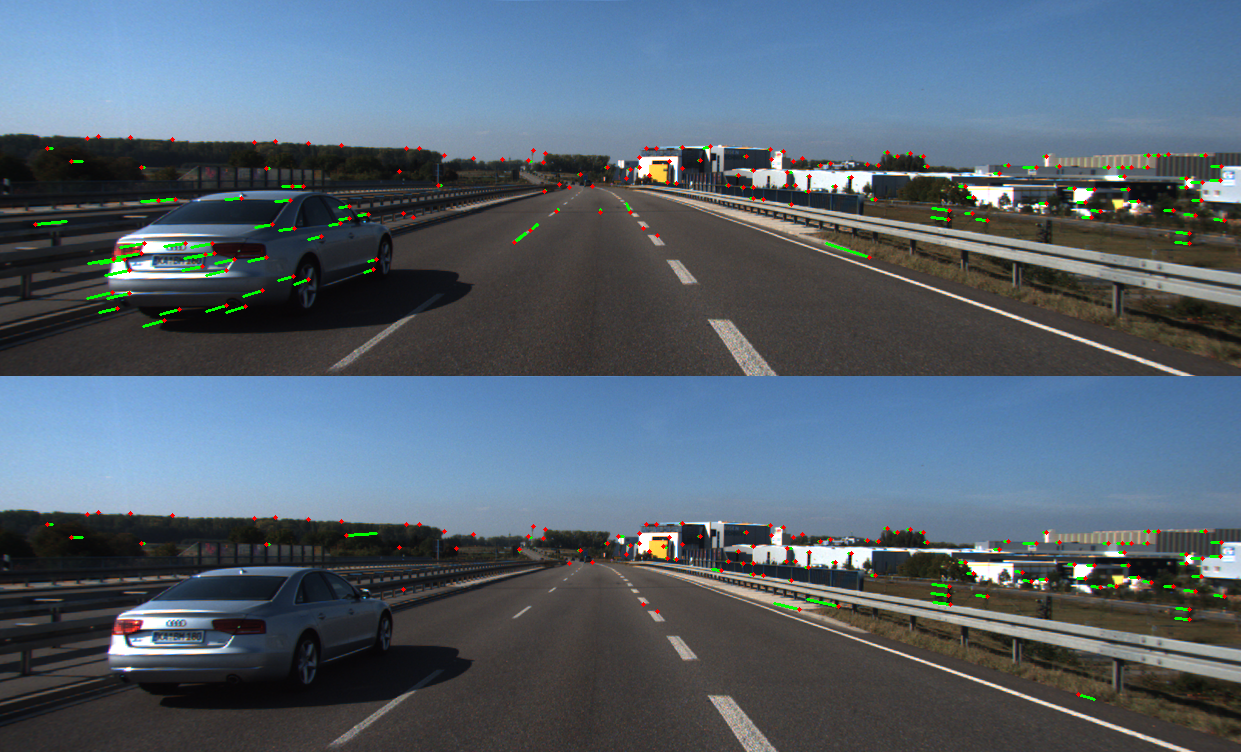}
		\caption{Sequence 01: Driving next to a car on a highway.
            Top: All keypoints.
            Bottom: Keypoints on car removed via its bounding box.}
		\label{fig:two images}
	\end{center}
\end{figure}
\begin{figure}[t!]
	\begin{center}
		\includegraphics[trim={15mm 1mm 15mm 1mm},clip,width=1.\linewidth]{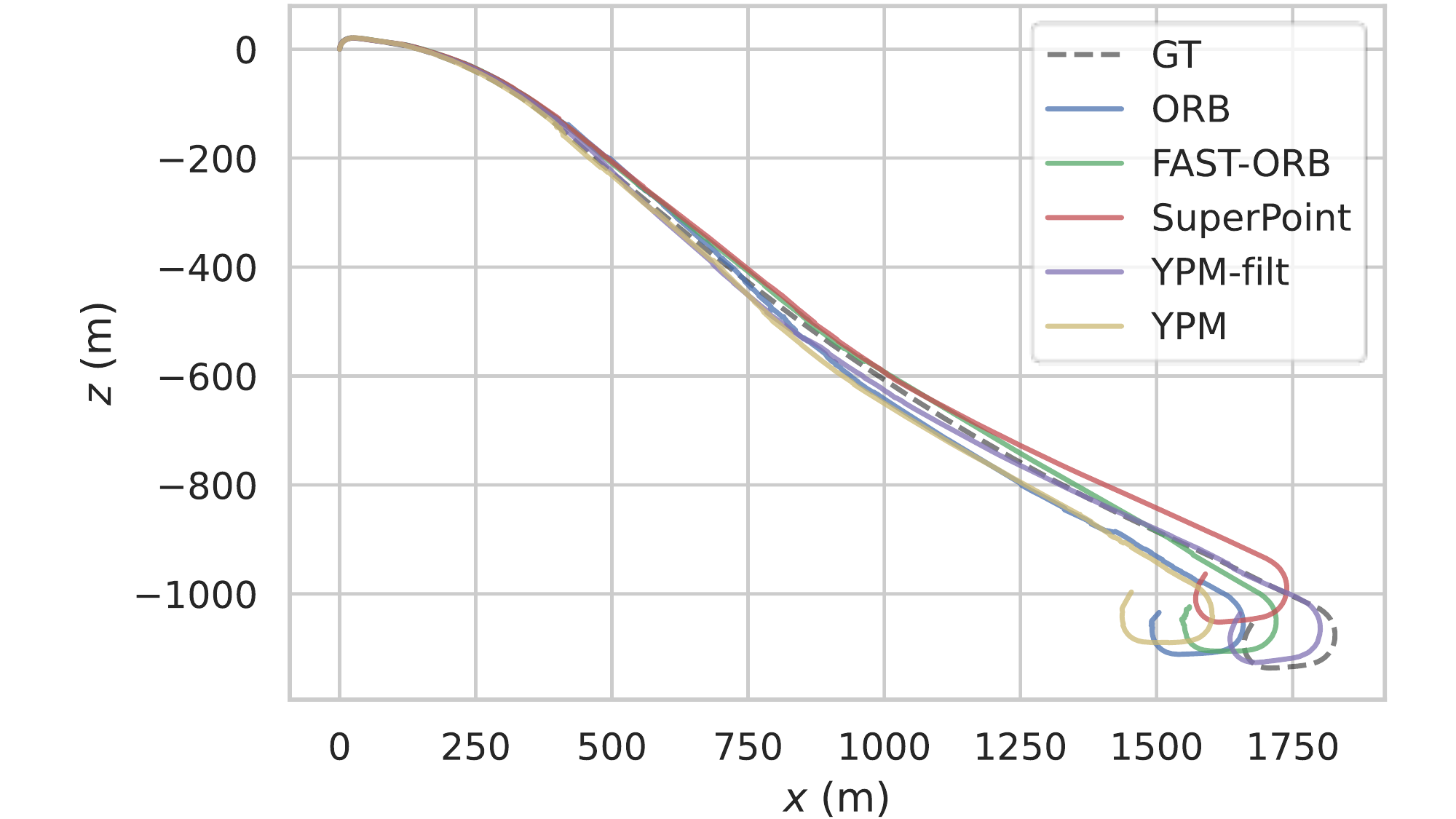}
		\caption{Sequence 01 trajectories of YOLOPointM (YPM) using filtered and unfiltered points and other keypoint detectors for comparison.}
		\label{fig:01 plot}
	\end{center}
\end{figure}
The efficacy of filtering out dynamic keypoints can best be demonstrated on sequence 01 as it features a drive on a highway strip where feature points are sparse and repetitive, i.e., hard to match.
Consequently, points detected on passing vehicles are no longer outweighed by static points (fig. \ref{fig:two images}), causing significant pose estimation error.
Fig. \ref{fig:01 plot} shows some trajectories including YOLOPointM with and without filtered points.
Without filtering, YOLOPoint performs comparatively poorly on this sequence due to it detecting many points on passing vehicles.
Finally, although all points on vehicles are removed, regardless of whether or not they are in motion, we do not find that this has a significant negative impact on pose estimation accuracy in scenes featuring many parked cars.

\section{Conclusion and Future Work}\label{Conclusion and Future Work}
In this work, we propose YOLOPoint, a convolutional neural network that jointly predicts keypoints, their respective descriptors and object bounding boxes fast and efficiently in a single forward pass, making it particularly suitable for real-time applications such as autonomous driving.
Our tests show that the keypoint detector heads that use a Darknet-like architecture perform similarly to SuperPoint in matching and repeatability tasks featuring strong view-point and illumination changes. On the KITTI visual odometry dataset we compare our model's pose estimation performance to SuperPoint and some classical methods.
Using predicted bounding boxes to filter out non-static keypoints, YOLOPoint shows the best accuracy to speed trade-off of all tested methods.

Future work will concentrate on incorporating YOLOPoint into our SLAM framework \cite{bib:beer2022itsc} by using keypoints and static objects as landmarks and increasing the robustness of object tracking \cite{tas:reich2022iros} by matching keypoints.

\bibliographystyle{splncs04}
\bibliography{anba}
\end{document}